\documentclass[letterpaper, 10 pt, conference]{ieeeconf}  

\IEEEoverridecommandlockouts                              

\usepackage{graphics} 
\usepackage{epsfig} 
\usepackage{mathptmx} 
\usepackage{times} 
\usepackage{amsmath} 
\usepackage{amssymb}  

\usepackage{romannum}
\usepackage[english]{babel}

\title{\LARGE \bf
Detecting Worker Attention Lapses in Human-Robot Interaction: An Eye Tracking and Multimodal Sensing Study
}

\author{Zhuangzhuang Dai$^{1}$, Jinha Park$^{2}$, Aleksandra Kaszowska$^{3}$, and Chen Li$^{4}$
\thanks{$^{1}$Z. Dai is with Dept. of Computer Science, Engineering and Physical Science, School of Engineering and Applied Science, Aston University, B4~7ET, Birmingham, United Kingdom
        {\tt\small z.dai1@aston.ac.uk}}%
\thanks{$^{2}$J. Park is with the SDU Robotics, The Maersk Mc Kinney Moller Institute, University of Southern Denmark, Campusvej 55, 5230 Odense, Denmark
        {\tt\small jipa@mmmi.sdu.dk}}%
\thanks{$^{3}$A. Kaszowska is with the Department of Electronic Systems, Aalborg University, Fredrik Bajers Vej 7, B5-215, 9220, Aalborg, Denmark
        {\tt\small kaszowska@es.aau.dk}}%
\thanks{$^{4}$C. Li is with the Department of Materials and Production, Aalborg University, Fibigerstræde 16, 9220, Aalborg, Denmark
        {\tt\small cl@mp.aau.dk}}%
}

\begin{document}

\maketitle
\thispagestyle{empty}
\pagestyle{empty}

\begin{abstract}

The advent of industrial robotics and autonomous systems endow human-robot collaboration in a massive scale. However, current industrial robots are restrained in co-working with human in close proximity due to inability of interpreting human agents' attention. Human attention study is non-trivial since it involves multiple aspects of the mind: perception, memory, problem solving, and consciousness. Human attention lapses are particularly problematic and potentially catastrophic in industrial workplace, from assembling electronics to operating machines. Attention is indeed complex and cannot be easily measured with single-modality sensors. Eye state, head pose, posture, and manifold environment stimulus could all play a part in attention lapses. To this end, we propose a pipeline to annotate multimodal dataset of human attention tracking, including eye tracking, fixation detection, third-person surveillance camera, and sound. We produce a pilot dataset containing two fully annotated phone assembly sequences in a realistic manufacturing environment. We evaluate existing fatigue and drowsiness prediction methods for attention lapse detection. Experimental results show that human attention lapses in production scenarios are more subtle and imperceptible than well-studied fatigue and drowsiness.

\end{abstract}


\section{INTRODUCTION}


Human attention is a cognitive process that involves a combination of physiological, psychological, and environmental attributes. Human attention lapses, characterized by the loss of focus or distraction, can significantly compromise task performance and bring about hazards for workers. Costs can be unbearably high when failing to detect attention lapses in circumstances such as assembling electronics and operating machines. In particular, the detrimental impact of attention lapses can result in errors, safety violations, and incurring significant impedance for industry to deploy autonomous robotic systems in context of Industrial 5.0.

Attention is a state in which one's cognitive resources are focused on certain aspects of the environment, and the operator is ready to respond to environmental stimuli. It can be inferred via physiological and behavioral indicators but not a physical entity to be measured directly. Attention lapses manifest widely across individuals due to different cognitive abilities, emotional states, and environment variations. These altogether make attention lapses difficult to monitor, nor to measure them accurately.

   \begin{figure}[tpb]
      \centering
      \includegraphics[width=0.9\linewidth]{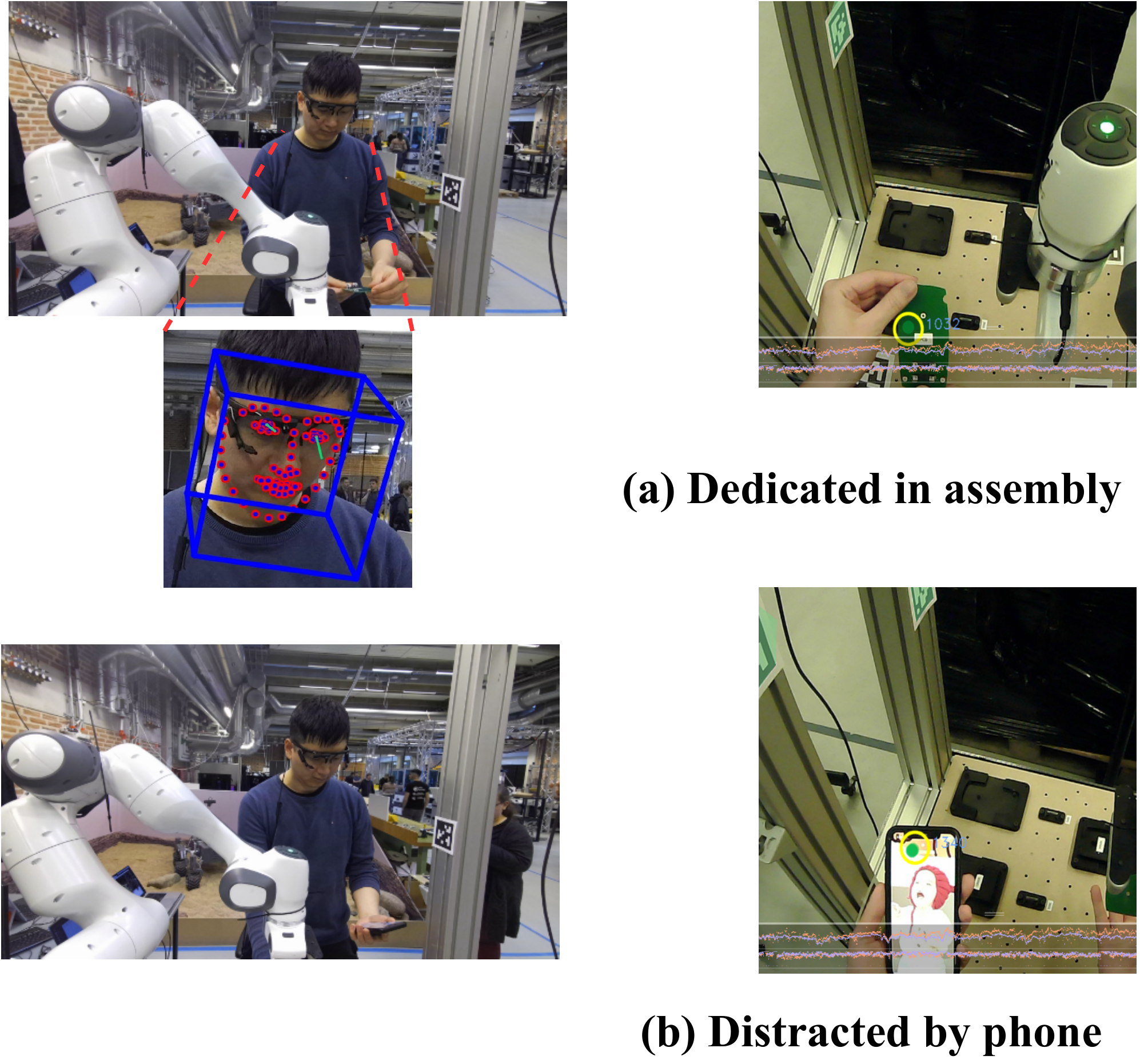}
      \caption{An example of distraction (looking at a mobile device) at workplace. Since the worker retains similar gaze, head pose, and posture in (a) dedicated and (b) distracted states, it is difficult to perceive such subtle difference without eye tracking. Vision-based detector, e.g., OpenFace~\cite{openface2016} (mid row), fails in contrasting attention lapses to normal states.}
      \label{fig:distract}
   \end{figure}

In order to track attention, our intuition is that many sensor modalities should work together to provide useful information. Given the \textit{multimodal} nature of attention, it can manifest in different ways, such as eye blinking, looking away, moving head or arm, speaking, or signs of drowsiness~\cite{humanODeyetrack2021}. In light of thriving multi-modality sensing systems, the combination of data on gaze, head pose, and audio can provide rich information for human attention and attention lapse study. Measuring attention lapses multimodally and automatically underlies a key aspect of promoting perception systems of social robots as well as improving worker safety. Social robots will benefit tremendously from better comprehending human's attention and disengagement, so as to make more appropriate interactions.


A multimodal sensing approach to attention lapse detection could assist in understanding the mechanisms of attention and developing effective interventions to improve human-robot collaboration. As can be seen from Fig. \ref{fig:distract}, a worker who looks in a correct direction with normal head pose and facial expression could be either engaged in working or distracted. Subtle differences in the worker's focus make recognizing such attention lapse a challenge. Without eye tracking to reveal human's cognitive state it is almost impossible to approach effective attention lapse detection. Nonetheless, relying solely on gaze modality will compromise reliability when there is occlusion of pupil or visual degradation.


To this end, we first curated a dataset of human operators conducting assembly task in a realistic production environment. A Franka robot played an assistive role in handing over components to the human agent~\cite{chen_howcanihelpu}. A Pupil Core eye tracker and a Microsoft Azure Kinect RGB-D camera with microphone are used for data collection. This setup allows multimodally and unobtrusively tracking the human operator's gaze, egocentric world view, as well as head pose, body posture, and environmental stimuli from a third-person perspective~\cite{dai_odom2022}. In order to streamline data annotation and pre-processing, we developed a pipeline to label distracted attention states frame-by-frame and visualizing the timelines. We also set up Apriltags~\cite{apriltag3} in the environment for head pose tracking. We used open-source tools~\cite{dai_deepaoanet2022,pupil2014,Ruiz_2018_CVPR_Workshops} to perform eye state classification (incl. fixation and blink) and fixated object detection. This pilot dataset has been annotated frame-by-frame to mark onsets of human attention lapses and in-operation states.


We also investigated the effect of applying existing fatigue and drowsiness detection methods~\cite{headgaze2014,perclos1998,blinkfrequency2019} on the pilot dataset. These methods have been widely used in driving safety domain. Unfortunately, they not only fall short in robustness for relying on single modality, but are hardly usable in real-world industrial settings where attention lapses take manifold forms. The results reveal a remarkable gap in accurate human attention lapse detection to allow safe and trustworthy HRI for industry.

In this work, our contribution is threefold: (1) We collected two pilot data sequences of multimodal human attention data in realistic industrial environments for benchmarking of human attention lapses detection; (2) We developed a pipeline for labelling multimodal sensory data, namely eye tracking, surveillance camera, and auditory input; (3) We experimented with existing fatigue and drowsiness prediction methods and identified a significant gap of knowledge in human attention lapse detection.

The rest of this paper is organized as follows. Section \Romannum{2} reviews existing literature around attention lapse detection datasets, methods, and applications. Our proposed multimodal annotation pipeline and benchmarking datasets are expanded in details in Section \Romannum{3}. Evaluation outcomes with existing methods are reported in Section \Romannum{4}. Section \Romannum{5} summarizes future work.

\section{BACKGROUND}
\label{sec:background}

\subsection{Human Attention Lapse Study}

Driving fatigue and drowsiness detection have been actively studied for its oriented application in driver safety domain. PERCLOS~\cite{perclos1998} is one of the most commonly used metrics for driving fatigue evaluation. An eye state tracking approach to PERCLOS is proposed in \cite{DrivingFatigue2015} to estimate eyelid closure ratio with respect to fatigue level. Although eye closure is agreed to be directly correlated to fatigue, human attention lapse can be quite polymorphic. Driver's eye looking away from the normal direction could be a sign of distraction or fatigue. Head pose~\cite{openface2016,Ruiz_2018_CVPR_Workshops} which is often associated with eye gaze may also indicate attention shifting.

Human attention lapses are of far more complex causes than fatigue and drowsiness. Indeed, a person may be mind wandering whilst having eye fixation at the right direction and objects. The fixation state of eyes may not directly indicate \textit{focused} or \textit{attention lapse} but the states of eye are closely correlated to human attention in general~\cite{AssembleCamp2021}. According to human cognition and recognition studies \cite{eyetrackingbook}, the following gaze patterns are strong indicators of attention lapse: long fixation on a single point, looking away, slow or irregular saccades, eye closure, and a high rate of blinking. Attention lapse is related to environmental distractors (auditory and visual), cumulative fatigue doing repetitive work, internal cognitive state such as motivation. A single-modality approach to determine human attention detection often falls short in accuracy and reliability~\cite{fatigueview2023}.

Our hypothesis is that attention lapse is multimodal, that is, human's internal cognitive state, environmental stimuli, together with eye state, facial expression, and posture collectively formulate trustworthy attention lapse detection. Existing driving fatigue detectors cannot be generalized for diverse applications~\cite{fatigueview2023,driverfatiguereview2022} in the sense that a driver sits in a confined position with a relatively simple but effective metric of ``looking ahead''. In this paper, we look into an open and more challenging scenario of assembly tasks in manufacturing. 

\subsection{Existing Datasets and Methods}

In the domain of driver fatigue research, detecting drowsiness and distraction using multimodal behavioural features~\cite{drowsyHOGSVM2015} allows applicable solutions to enhance driver safety. Transferring this idea to HRI could mean a big step forward for workers' safety in automated production lines~\cite{dai_ble2020}. According to ~\cite{driverfatiguereview2022}, it is found the combination of eye feature (gaze, blinking, and fixation states) and head pose achieve reliable estimations. Methods based on single-modality turn out susceptible to occlusions and hard to generalize well. In \cite{headgaze2014}, the authors proposed handcrafted thresholds of $30^{\circ}$ nodding and 200-frame (about 6.7s) no change in gaze direction to detect drawsiness, which can never produce satisfactory reliability. Yawning makes an effective indicator of fatigues but becomes less useful in distraction detection~\cite{fatigueview2023}.

Static human attention tracking within a fixed plain (e.g., looking at a flat screen) has been well-studied, such as SNAG dataset \cite{SNAG2022} and human attention dataset in image captioning \cite{AttentionForCaption2019}. Recently more focuses have been drawn on studying human attention in action such as assembling a camping tent \cite{AssembleCamp2021} and driver monitoring~\cite{DrivingFatigue2015}.

Modern eye trackers achieve fairly high accuracy, precision, and effective frequency~\cite{EyeTrackers2022} even wearers performing substantial movements. Nevertheless, eye tracking in unrestrained settings, such as lost track of pupil and long eye closure, will inevitably affect the data quality in accuracy and precision~\cite{unrestrained2018}. Furthermore, a robust eye state tracking entails high-fidelity visual monitoring from third-person views~\cite{dai_egocap2023}. This deteriorates credibility of predicting attention by eye tracking only.

It has been widely acknowledged environmental noises, such as those in industrial workplace, have a profound impact on human cognition~\cite{noisepollution2022}. Auditory stimuli are also common external stimuli that draw human attention and arouse distraction~\cite{NoiseAttention2019}. We are thus motivated to collect a multimodal human attention tracking dataset tailored for cooperation between human agents and robot manipulators in industrial settings. As aforementioned, human's eye states, head pose, body posture, and sound stimuli shall all be synchronized and recorded. The following sections expands our data collection campaign, annotation pipeline, and evaluation results.


\section{METHOD}

\begin{figure}[t]
      \centering
      \includegraphics[width=0.98\linewidth]{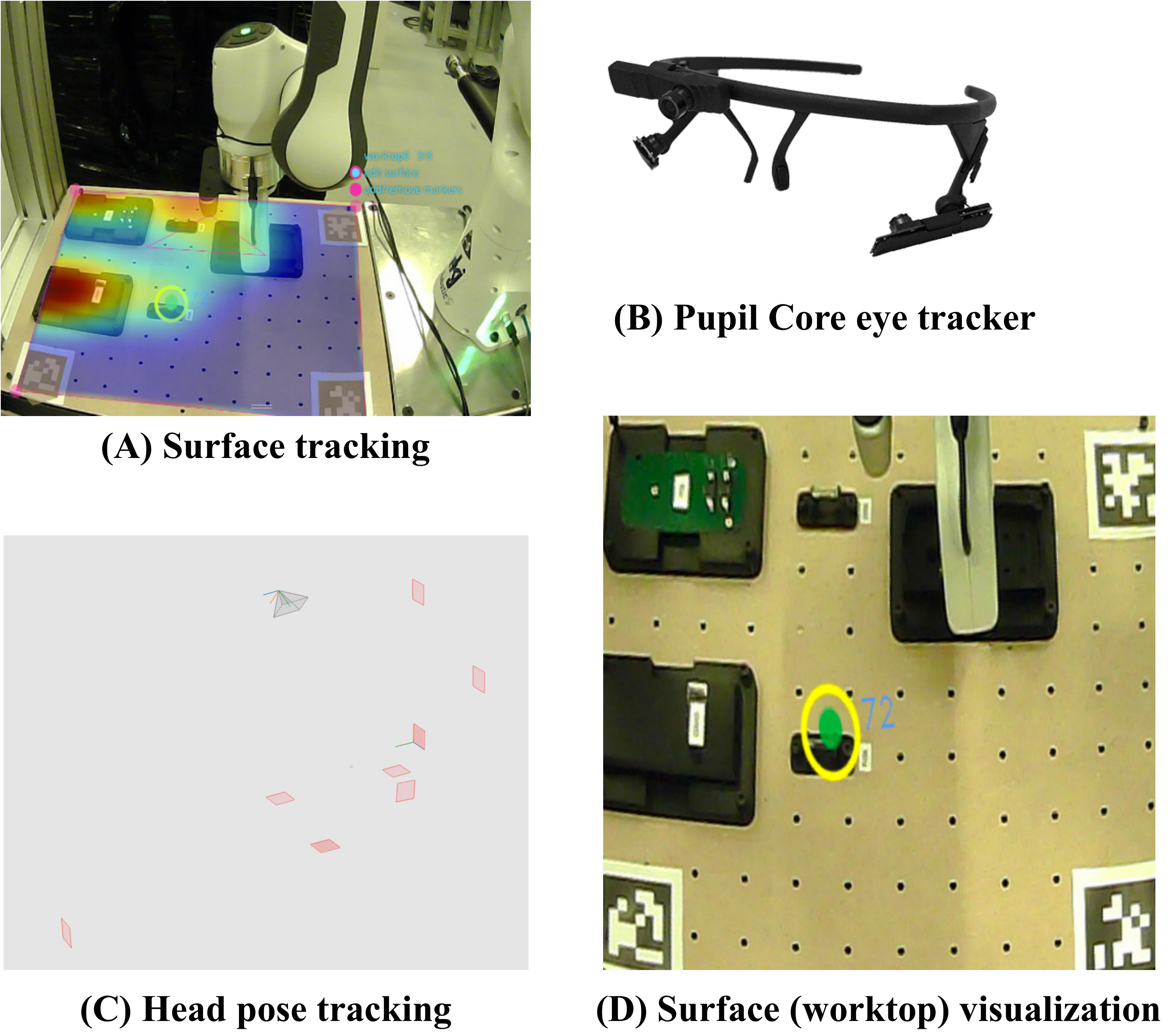}
      \caption{Visualization of the data collection environment and annotations. (A)(D) show the assembly worktop defined by QR codes and its 2D projection with gaze in post-hoc processing. Green dots are gaze projections; yellow circle marks the fixation position. (B) shows the Pupil Core eye tracker. Head pose tracking results are shown in (C).}
      \label{fig:setup}
\end{figure}

We collected a pilot dataset of a human worker doing a dummy phone assembly task with the aid of a pre-programmed robot manipulator. Multi-modality sensors, namely an RGB-D camera with microphone (Azure Kinect DK) and an eye tracker with egocentric world camera (Pupil Core), are deployed with multiple artificial distractors injected during the experiment process. An automated annotation pipeline is proposed to streamline the labelling of multiple primitives for human attention lapse research. We also utilized pre-trained object detectors~\cite{fasterrcnn2015} for fixated object recognition and pre-trained head pose detector~\cite{Ruiz_2018_CVPR_Workshops} for head pose calibration.

\subsection{Pilot Dataset}

   \begin{figure}[tp]
      \centering
      \includegraphics[width=0.99\linewidth]{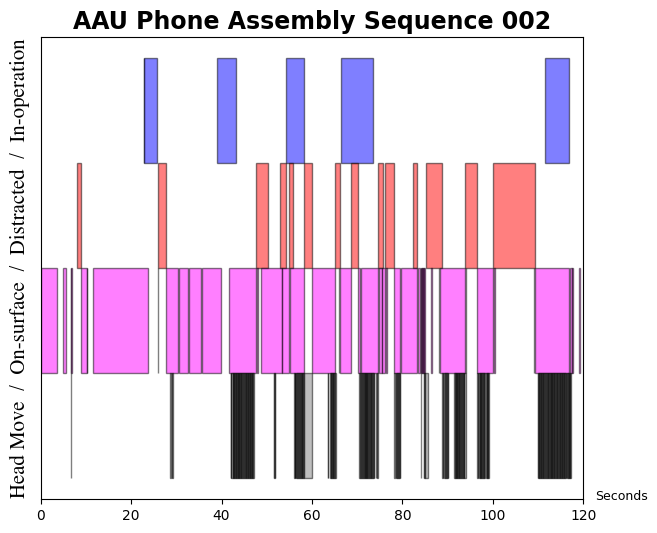}
      \caption{Annotated in-operation and distracted states in phone assembly sequence 002 compared with classic estimation methods. Blue bars represent onsets of the worker's hands-on operation which implies greater hazards upon robotic intervention. Red bars represent onsets when the worker is distracted. Magenta bars indicate eye fixations within the predefined surface (i.e., \textit{worktop}). Grey bars mark head movements with a high angular velocity. These events are annotated against temporal UTC timestamps of millisecond (msec) precision.}
      \label{fig:sequence002}
   \end{figure}

\begin{table*}[t]
\centering
\caption[labels]{Annotations in the pilot dataset}
\label{table:labels}
\begin{tabular}{|c|c|c|}
\hline
\textbf{Label} & \textbf{Type} & \textbf{Units}\\
\hline
Eye state & \textit{fixation/blink/saccade} & Classes and msec \\
\hline
Gaze & 2D position & Relative coordinates\\
\hline
Fixation & 2D position & Relative coordinates\\
\hline
Fixation object & Detected object class & Classes\\
\hline
Head pose & 6 degree-of-freedom pose & Pose\\
\hline
Fixation in AOI & 2D position and whether fixation is in AOI & Relative coordinates and True/False\\
\hline
Distractor & \textit{noise/motion/phone} and onset timestamps & Classes and msec\\
\hline
Distracted state & Start and end timestamps & msec\\
\hline
In-operation state & Start and end timestamps & msec\\
\hline
\end{tabular}
\end{table*}

The data collection was conducted in Aalborg University 5G Smart Production Lab~\cite{smartlab2017}. It provides a realistic manufacturing environment to present real-world industrial scenarios. A Franka robot has been pre-programmed to harvest five components of a dummy phone, shown in Fig. \ref{fig:setup}(D), and hand them over by releasing the components ahead of human operator's right hand position. The release takes place at slightly varying locations which requires human's attention to catch the components. Note these five dummy components (one front cover, one back cover, one printed circuit board, and two fuses) are made of plastic and would cause no harm if caught unsuccessfully. A software-based stop command and an emergency button are placed next to the human participant for safety. A human experimenter was asked to perform phone assembly tasks while multimodal data of gaze, vision, speech, and environmental stimuli from multiple angles were collected. During the task, the dummy phone parts placed on a flat worktop will be picked up by the Franka robot and the human participant was expected to put these parts together, as is shown in Fig. \ref{fig:setup}(A). 

We experimented with a Pupil Core eye tracker as shown in Fig. \ref{fig:setup}(B). Pupil Core offers binocular eye tracking at 200Hz (highest among mass marketed wearable and mobile products) and a scene camera to shoot world view of $155^{\circ}/85^{\circ}$ horizontal/vertical field of view with configurable resolution and frame rate~\cite{pupil2014}. The eye cameras take infrared illuminated images and detects the pupil area with a robust and efficient algorithm and multiple noise filters (an average gaze detection accuracy of $0.6^{\circ}$ of visual angle and a processing latency of 45ms). Pupil Core stands out for its light-weight design, minimized visual obstruction, and ability to accommodate varying facial geometries in comparison to SMI ETG2 60 or Tobii G-series~\cite{EyeTrackers2022}. More importantly, the open source software of Pupil Core allows customization in data collection, processing, and labelling processes. Data streaming, blink detection, and head pose tracking modules can be integrated as plugins at ease. Eye tracking endows the ability of detecting introverted human attention in a direct and non-obstructive means~\cite{apriltag3}.

A Microsoft Azure Kinect DK with microphone array was used for a third-person perspective recording. This kit contains an RGB-D camera with infrared sensing, 7-microphone array, and integrated body pose tracking backends. We used a Lenovo laptop with an Intel i7 processor and Ubuntu 18.04 operating system installed for data collection. All sensors are synchronized by the host laptop's local clock.

QR codes are commonly used for head pose estimation in eye tracking practices. Following the method proposed in \cite{fastQR2021}, we deployed multiple Apriltags~\cite{apriltag3} in the environment and utilized the post-hoc pose tracking algorithm of the Pupil Core software for head pose estimation. We deployed eight Apriltags from the \textit{tag36h11} family next to the assembly worktop as shown in Fig. \ref{fig:setup}(C,D). Three of them are attached to the flat surface of the worktop to allow surface abstraction and gaze tracking in Area of Interest (AOI).

\subsection{Annotation Pipiline}

To achieve multimodal attention detection, we developed a data annotation pipeline for efficient processing of the multimodal data. First, we recognized a list of significant primitives in human attention study as detailed in Table \ref{table:labels}. Specifically, the eye state, gaze position, and fixation position can be extracted from Pupil Core's software system outputs but require careful calibration as errors are common in default detectors. The fixation object is automatically annotated by a pre-trained object detector~\cite{fasterrcnn2015} with manual corrections. Head pose and surface tracking are derived with the aid of eight Apriltags deployed next to the assembly platform. Of our greatest interest, the distracted states and in-operation states are manually labelled frame-to-frame to identify subtle attention lapses. An example of the annotated sequence 002 is shown in Fig. \ref{fig:sequence002}.

Note ensuring ad-hoc occurrence of real-world distractors in an assembly task is critical for attention lapse study. We curated two pilot data sequences of a human participant assembling a dummy phone in a realistic production laboratory with manufacturing background noises, moving MiR robots, and workers around. According to K. Holmqvist and R. Andersson~\cite{eyetrackingbook}, the following distractors in industrial settings are identified:

\begin{itemize}

\item Noise (background/sudden)
\item Co-workers (talk to ``me''/talk to each other)
\item Malfunction (user mistake/equipment)
\item Fatigue (drowsiness/mind wandering)
\item Visual stimuli (environment/moving objects)
\item Multitasking
\item Motivation (task difficulty/repetitiveness/reward)

\end{itemize}

\begin{figure*}[!tph]
    \centering
    \includegraphics[width=0.86\linewidth]{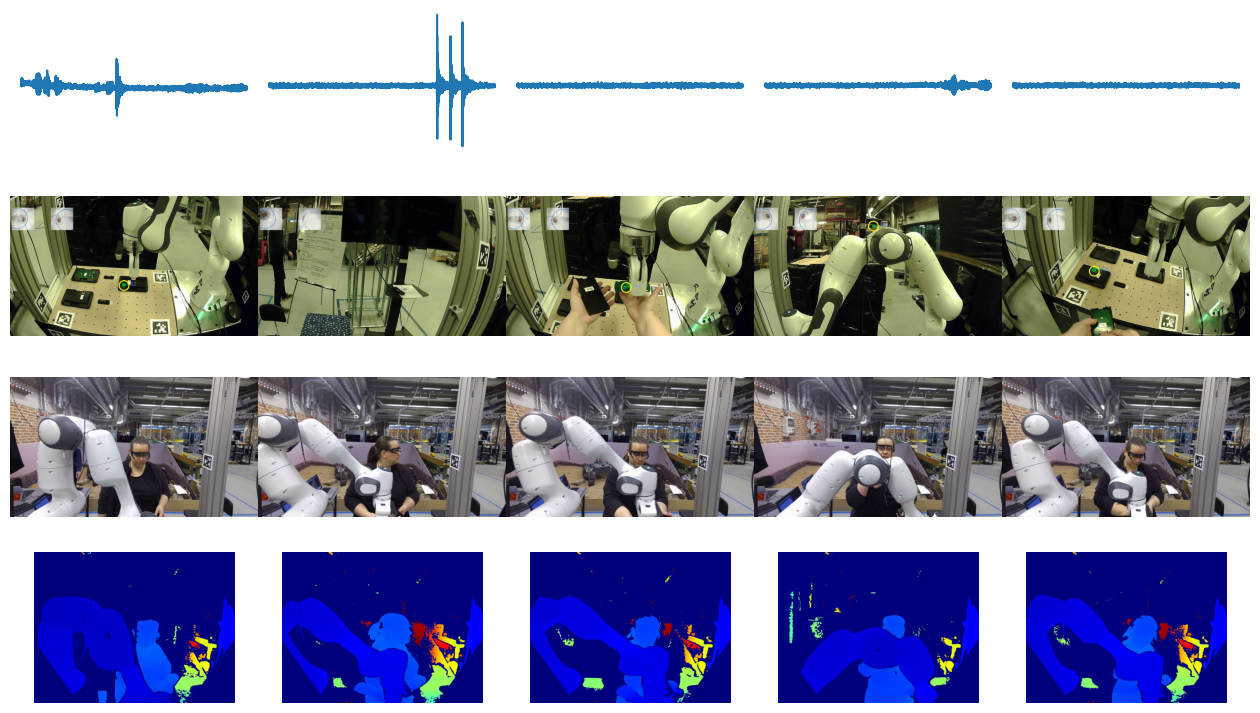}
    \caption{Visualization of audio distractors (1st row), eye tracker's world view (2nd row), surveillance RGB scene (3rd row), and depth scene (4th row). The second frame (column) shows the worker being distracted by hand claps to her left. The fourth frame (column) illustrates distraction of talking to co-workers.}
    \label{fig:view}
\end{figure*}

During the pilot data collection, we introduced three of the above distractors due to constraints of the environment and safety concerns: clapping hands (sudden noise), walking co-workers at the scene (visual distraction), and looking at a personal smart phone (mind wandering). In the first clip (Sequence 002), we asked the experimenter to be open-minded to attention lapses and look along distractors. In the second clip (Sequence 002), the experimenter was asked to stay as concentrated as possible in which the only distractor is \textit{mind wandering}. The onset of these events have been post-hoc annotated by checking multiple camera views and auditory recordings as shown in Fig. \ref{fig:view}.

A FasterRCNN-based object detector~\cite{fasterrcnn2015} is leveraged to perform post-hoc fixation object recognition. We applied the object detector on every frame and used fixation point as a filter to only annotate object of attentive interest. If the fixation is more than 15-pixel away from an object's bound box the detected object will be omitted as the fixation lands far apart from the object. Errors are inevitable using such automated annotations. We further validated and corrected wrong labels frame-by-frame.



\section{EVALUATION}

We experimented with existing single-modality methods on the pilot dataset. Classic fatigue and drowsiness detection modalities include ratio of eye closure state (PERCLOS)~\cite{perclos1998}, blink frequency~\cite{blinkfrequency2019}, gaze fixation state~\cite{eyetrackingbook}, and head pose tracking~\cite{headgaze2014}. We examined their performance in determining attention lapse upon phone assembly sequences 002 and 003.

\begin{table}[t]
\centering
\caption[eval]{Detection Results on Phone Assembly Sequence 002.}
\label{table:eval002}
\begin{tabular}{|c|c|c|c|}
\hline
\textbf{Method} & \textbf{Precision} & \textbf{Recall} & \textbf{F1 Score}\\
\hline
Eye closure period (PERCLOS) & $35.71\%$ & $37.42\%$ & $0.37$\\
\hline
Eye blink frequency & $19.55\%$ & $13.16\%$ & $0.16$\\
\hline
Fixation within AOI & $8.70\%$ & $19.31\%$ & $0.12$\\
\hline
Head movement velocity & $13.11\%$ & $9.49\%$ & $0.11$ \\
\hline
\end{tabular}
\end{table}

\begin{table}[t]
\centering
\caption[eval]{Detection Results on Phone Assembly Sequence 003.}
\label{table:eval003}
\begin{tabular}{|c|c|c|c|}
\hline
\textbf{Method} & \textbf{Precision} & \textbf{Recall} & \textbf{F1 Score}\\
\hline
Eye closure period (PERCLOS) & $0$ & $0$ & N/A\\
\hline
Eye blink frequency & $4.24\%$ & $100\%$ & $0.08$\\
\hline
Fixation within AOI & $3.15\%$ & $52.22\%$ & $0.06$\\
\hline
Head movement velocity & $0.85\%$ & $24.39\%$ & $0.02$ \\
\hline
\end{tabular}
\end{table}

PERCLOS refers to the percentage of duration of closed-eye state within a given time interval (30 seconds or 1 minute). For a fair comparison, we use EM criteria, which represents 50\% of eye closure rate, upon a 30s sliding window to metric attention lapse on the phone assembly data sequences. According to \cite{blinkfrequency2019}, eye blink frequency drops below 10 per minute when sleepy. We adopted this threshold for blink frequency based drowsiness estimation in a 30s sliding window.

Workers are expected to pay attention to an AOI, such as a worktop, when focusing on specific tasks. This has been widely accepted as an important attribute in attention detection. We defined a worktop area using Pupil Core's software as shown in Fig. \ref{fig:setup}(A). The entry and exits from this AOI can be calculated as shown in Fig. \ref{fig:sequence002} ``On-surface''. I. Choi and Y. Kim~\cite{headgaze2014} contended nodding over $30^{\circ}$ in a short time implies drowsiness. We extend this to head rotation around any axis for distraction detection. We specified the short time equals $2$ seconds. Given densely deployed QR codes, pose-hoc head poses can be reconstructed as seen in Fig. \ref{fig:setup}(C). Thus, we obtained distraction states as shown in Fig. \ref{fig:sequence002} ``Head Move''.

Overall, poor precision, recall, and F1 scores are seen from pilot data sequences as shown in Tables \ref{table:eval002} and \ref{table:eval003}. PERCLOS turns out the most reliable attention lapse detector when open-minded to distraction. Yet its precision and recall are far from satisfactory. Other single-modality methods demonstrate extremely low coherence with true distraction states. In Sequence 003 where experimenter was mentally prepared, all fatigue and drowsiness methods fail. These back up our judgement that attention lapse is multimodal and complex, suggesting simple paradigms of monitoring eye blinks, fixation, or head pose cannot be trusted. Additionally, single or few sensors for detection may suffer in accuracy when occlusion or data discrepancy occur. For instance, we found some \textit{long-lasting} blinks, i.e., over 5s, in Pupil Core's algorithm for blink detection. This may be due to losing tracks of pupils. We conclude from our pilot study that a multimodal and multi-sensory scheme for worker attention lapse detection becomes imperative in this research area.



\section{FUTURE WORK}

In this work, we addressed data primitives of a human attention lapse study and created a pilot dataset to evaluate existing detection methods. The pilot dataset collected falls short in size to support data-driven algorithms for attention lapse detection, such as SVM and deep neural networks. We plan to launch a large-scale data collection campaign with dozens of human participants accomplishing various assembly tasks. Current experimental setup supports one set of sensors on eye tracker and the other set in front of the workspace. We will expand it with more sets of sensors from multiple angles in the complete dataset.

We notice the frame rates of multimodal sensors differ. Eye cameras run at 200Hz, whereas, world camera and Azure Kinect DK run at 30Hz. Furthermore, there is an offset between sound track and RGB-D video. We will elaborate the data collection pipeline to achieve enhanced synchronisation. We also plan to integrate threads of automatic body posture tracking and hand gesture recognition into the annotation pipeline.






\section*{ACKNOWLEDGMENT}

This work was funded by 2022/23 Aston Pump Priming Scheme and AAU Bridging Project ``A Multimodal Attention Tracking In Human-robot Collaboration For Manufacturing Tasks.'' We thank the Aalborg 5G Smart Production Lab for supporting our data collection campaign.

\bibliographystyle{plain}
\bibliography{root}

\end{document}